%% file: main.tex
\documentclass[a4paper]{./styles/svproc}

\usepackage{url}
\usepackage{graphicx}
\usepackage{subcaption}
\usepackage{xcolor}
\usepackage{amsmath}

\usepackage[
backend=biber,
style=ieee,
sorting=ynt
]{biblatex}
\usepackage{wrapfig}

\date{21.05.2022}

\begin{document}

\mainmatter              
\title{Ball-and-socket joint pose estimation using magnetic field}
\titlerunning{Ball-and-socket joint pose estimation using magnetic field} 

\author{Tai Hoang\inst{1}, Alona Kharchenko\inst{2}, Simon Trendel\inst{2}, Rafael Hostettler\inst{2}}
\authorrunning{Tai Hoang et al.} 
%
\tocauthor{Tai Hoang}
\institute{Technical University of Munich, Munich, Germany,\\
    \email{t.hoang@tum.de}
    \and
    Devanthro – the Robody Company, Munich, Germany \\
}

\maketitle

\begin{abstract}
  \input{sections/00_Abstract}

\end{abstract}

\section{Introduction}
\label{sec:Introduction}
\input{sections/01_Introduction}

\section{Hardware}
\label{sec:Hardware}
\input{sections/02_Hardware}

\section{Methodologies}
\label{sec:Methodologies}
\input{sections/03_Methodologies}

\section{Experiments}
\label{sec:Experiments} 
\input{sections/04_Experiments}

\section{Conclusions}
\label{sec:Conclusion}
\input{sections/05_Conclusion}

\section*{Acknowledgements}
We would like to thank Maximilian Karl for the helpful discussions on DVBFs model and Jurgen Lippl for the 3D model of the ball-socket-joint.

\printbibliography

\end{document}

%% file: sections/00_Abstract.tex
Roboy 3.0 is an open-source tendon-driven humanoid robot that mimics the musculoskeletal system of the human body. Roboy 3.0 is being developed as a remote robotic body - or a robotic avatar - for humans to achieve remote physical presence. Artificial muscles and tendons allow it to closely resemble human morphology with 3-DoF neck, shoulders and wrists. Roboy 3.0’s 3-DoF joints are implemented as ball-and-socket joints. While industry provides a clear solution for 1-DoF joint pose sensing, it is not the case for the ball-and-socket joint type. In this paper we present a custom solution to estimate the pose of a ball-and-socket joint. We embed an array of magnets into the ball and an array of 3D magnetic sensors into the socket. We then, based on the changes in the magnetic field as the joint rotates, are able to estimate the orientation of the joint. We evaluate the performance of two neural network approaches using the LSTM and Bayesian-filter like DVBF. Results show that in order to achieve the same mean square error (MSE) DVBFs require significantly more time training and hyperparameter tuning compared to LSTMs, while DVBF cope with sensor noise better. Both methods are capable of real-time joint pose estimation at 37 Hz with MSE of around 0.03 rad for all three degrees of freedom combined. The LSTM model is deployed and used for joint pose estimation of Roboy 3.0's shoulder and neck joints. The software implementation and PCB designs are open-sourced under \url{https://github.com/Roboy/ball_and_socket_estimator}

%% file: sections/01_Introduction.tex
Classical rigid robots, which often consist of a chain of rigid and heavy links, pose a safety risk in unstructured environments. Soft robots, on the other hand, made primarily of lightweight materials, have a great potential to operate safely in any environment, especially during dynamic interaction with humans. Many soft mechanical designs have been proposed in recent years. Musculoskeletal robot \cite{roboy} or more specifically, Roboy 3.0, an open-source humanoid robot, is one of the few examples of the soft humanoid robot designs. It mimics the working principle of the human musculo-skeletal system and morphology. Instead of having an actuator directly in the joint like in classic rigid robots, Roboy 3.0 links are actuated by a set of artificial muscles and tendons - series elastic actuators.
\begin{wrapfigure}{r}{0.35\textwidth}
  \begin{center}
    \includegraphics[width=0.35\textwidth]{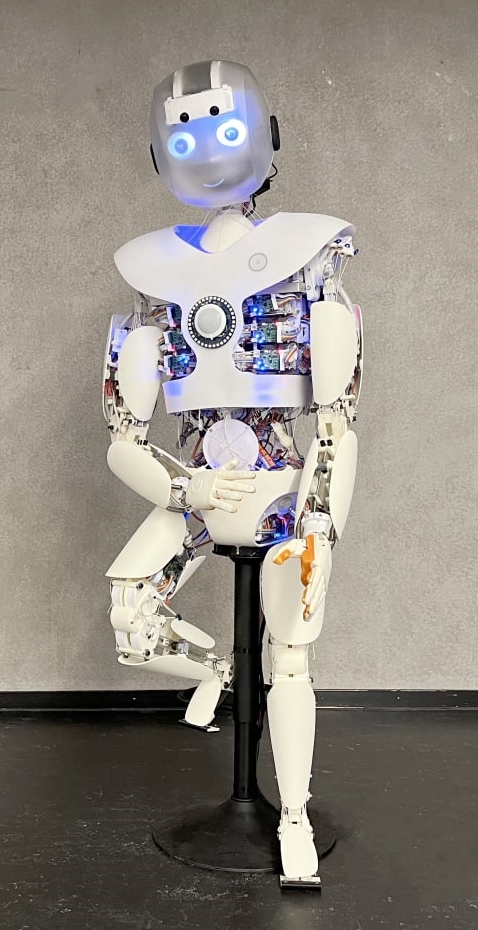}
  \end{center}
  \caption{Tendon-driven humanoid robot Roboy 3.0}
  \label{fig:roboy}
\end{wrapfigure}

Such design allows to passively store the energy in the muscles and makes the robot inherently compliant. These properties, combined with anthropomorphic morphology, turn to be beneficial in full-body motion mapping during teleoperation as well collaborative object manipulation and other close physical interactions with humans. However, achieving reliable, stable and precise joint and end-effector-workspace control for a tendon-driven humanoid \cite{cardsflow} is challenging due to difficulties in precise modelling of muscle co-actuation and tendon friction among other things. Thus, a high-accuracy robot state estimation is of crucial importance to enable the implementation of closed-loop control algorithms. 
There are two types of joint being used in the upper body of Roboy 3.0: 3-DoF and 1-DoF. The neck, shoulders, and wrists are 3-DoF joint and elbows are 1-DoF. Determining the joint position for 1-DoF joints is possible with high precision using off-the-shelf sensors. However, the same solution is not directly applicable for the 3-DoF joint case. We therefore, in this paper provide a solution on both hardware and software for this 3-DoF ball-socket joint.


Our idea is based on using the magnetic field of strong neodymium magnets to determine the orientation of the joint. This has long been a well known approach due to various reasons. First, it is a contact-free measurement, and way more robust to the environment compared to other contact-free sensors like optical \cite{Lee04} and vision-based \cite{Garner01}. Second, the measurement value can be obtained at high rates with a high accuracy. The sensor devices are also not expensive and have a sufficiently small size which makes it easy to integrate into our robot. However, obtaining the orientation directly from the magnetic value in closed-form is challenging due to their non-linearity relation. We therefore shift the focus on the data driven approach, or more particularly, learning a neural network to do the transformation between the magnetic sensors output and the joint's orientation. Recently, several researchers have successfully developed solutions based on this idea. Jungkuk \cite{Junguk17} built a full pipeline to obtain the orientation of a 2-DoF object with only a single sensor, Meier \cite{Phil19} developed a system that can determine the human hand-gesture based on the magnetic field. Magnetic field can also be used to determine the position of an indoor robot, which is an active research area for indoor localization problems \cite{Chiang20, Sasaki22}.

The magnetic sensor used in this paper is a three-dimensional Hall-effect sensor. This particular type of sensor has a critical issue: it is very noisy and since our ultimate goal is using the inferred orientation to close the control loop, it is very important to mitigate this issue, otherwise our robot could not operate safely and can easily damage itself and the environment. In this paper, we proposed to use two types of neural network that can deal with this problem: a recurrent-based neural network (LSTM) \cite{HochSchm97} and Deep Variational Bayes Filter (DVBF) \cite{karl2017deep, karl2017unsupervised}. We hypothesize that both of the approaches can smoothen out the output signal. While the former does this based on the historical data, the latter approach is designated to do this more explicitly with the integrated Bayesian filter.

%% file: sections/02_Hardware.tex
\begin{figure}[ht] 
  \centering
  \begin{subfigure}[b]{0.45\linewidth}
    \centering
    \includegraphics[width=\linewidth]{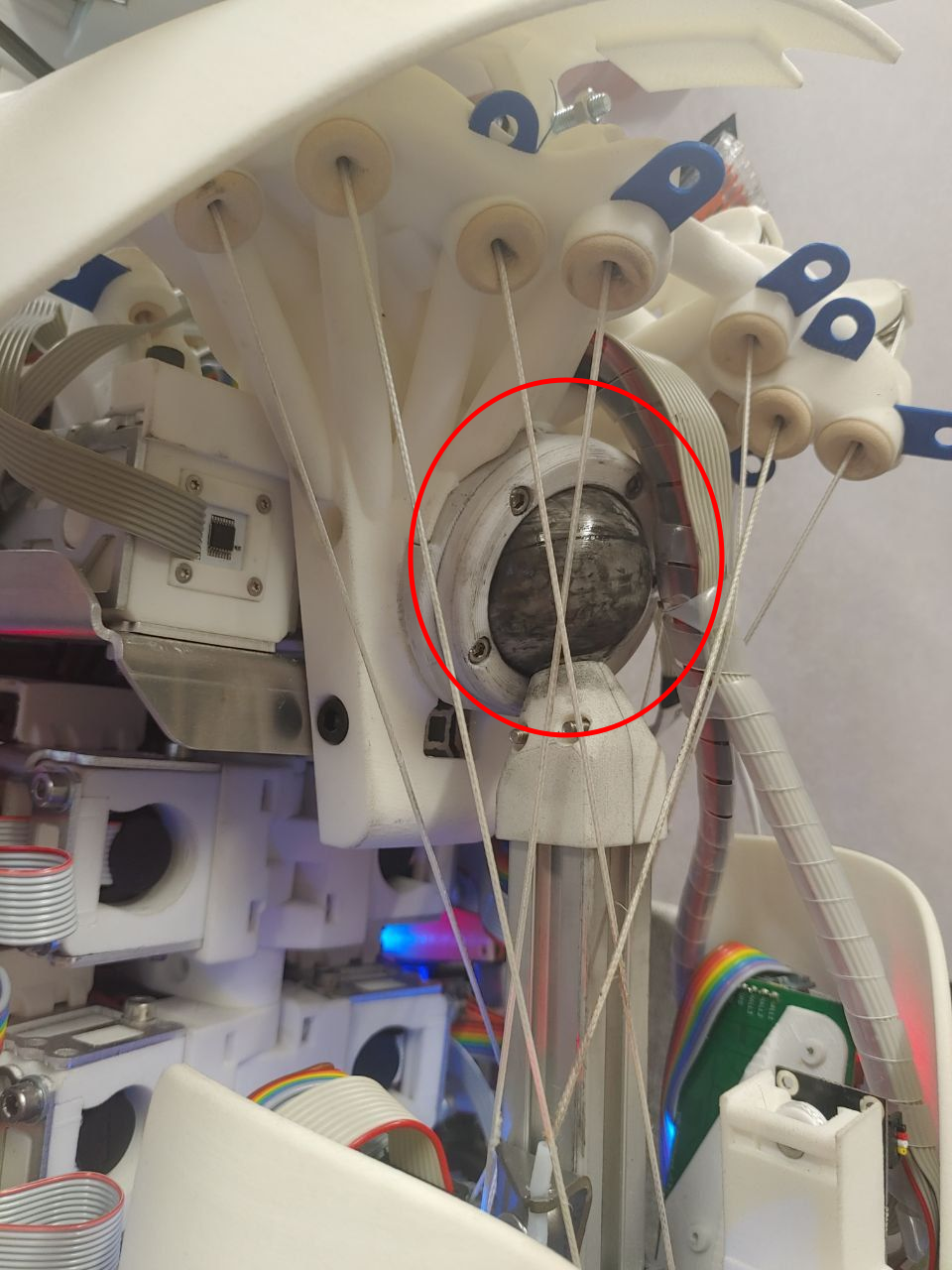}
    \caption{ball-in-socket joint} 
    \label{fig_hw:1} 
  \end{subfigure}
  \qquad
  \begin{subfigure}[b]{0.45\linewidth}
    \centering
    \includegraphics[width=\linewidth]{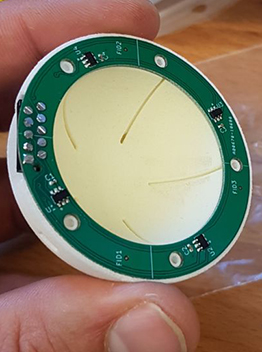}
    \caption{printed circuit board} 
    \label{fig_hw:2} 
  \end{subfigure} 

  \caption{Hardware design of Roboy 3.0 shoulder ball-and-socket joint}
  \label{fig_hw} 
\end{figure}

\begin{figure}[ht] 
  \centering
  \begin{subfigure}[b]{0.45\linewidth}
    \centering
    \includegraphics[width=\linewidth]{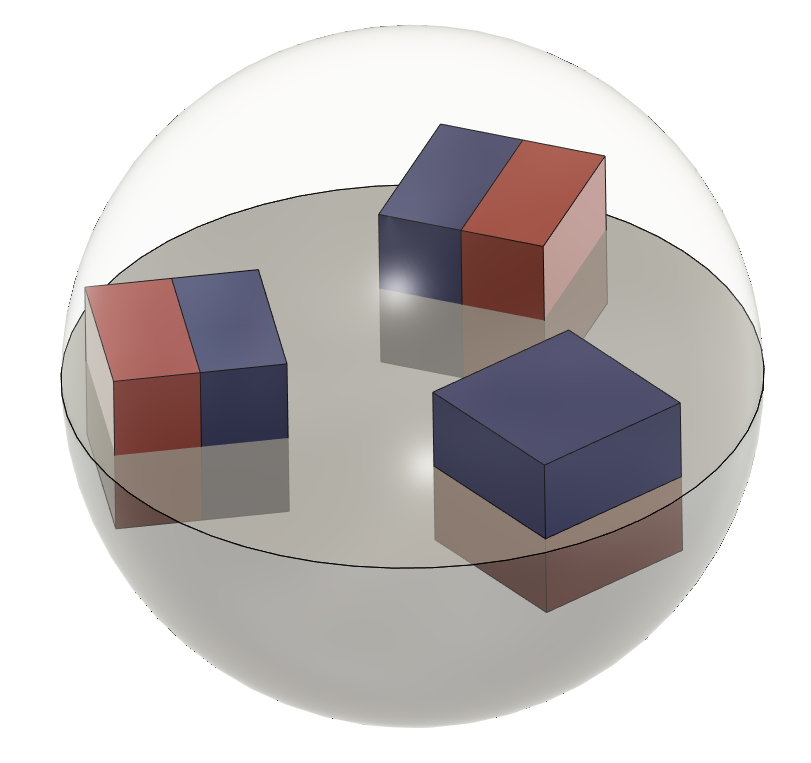}
    \caption{Side view} 
    \label{fig_hw:magnet_side} 
  \end{subfigure}
  \qquad
  \begin{subfigure}[b]{0.45\linewidth}
    \centering
    \includegraphics[width=\linewidth]{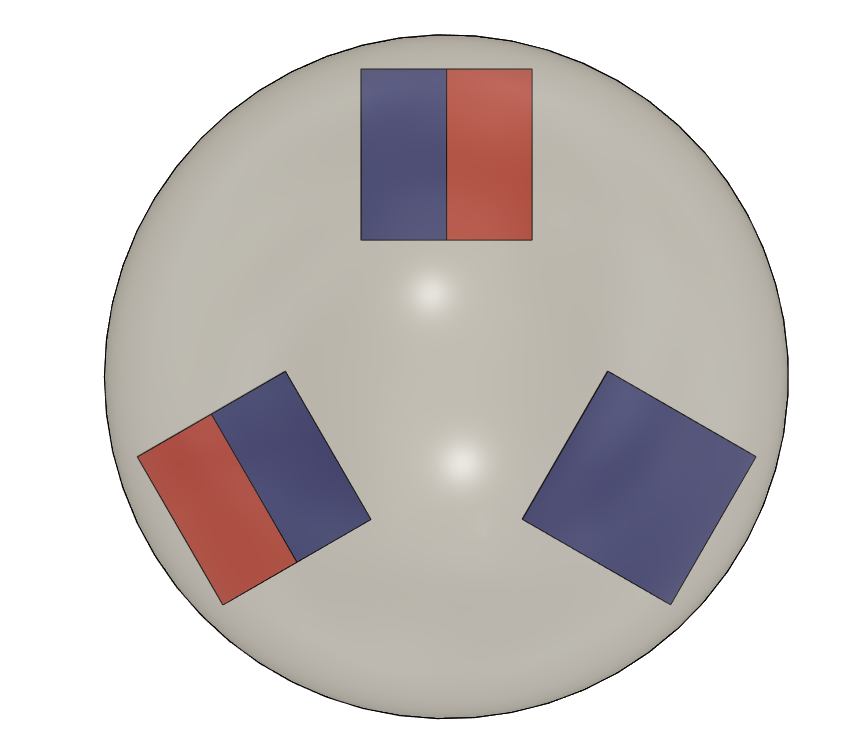}
    \caption{Top view} 
    \label{fig_hw:magnet_top} 
  \end{subfigure}
 
  \caption{Location of magnets inside the ball}
  \label{fig_hw:magnet} 
\end{figure}

\begin{figure}[ht] 
    \centering
    \includegraphics[width=0.6\linewidth]{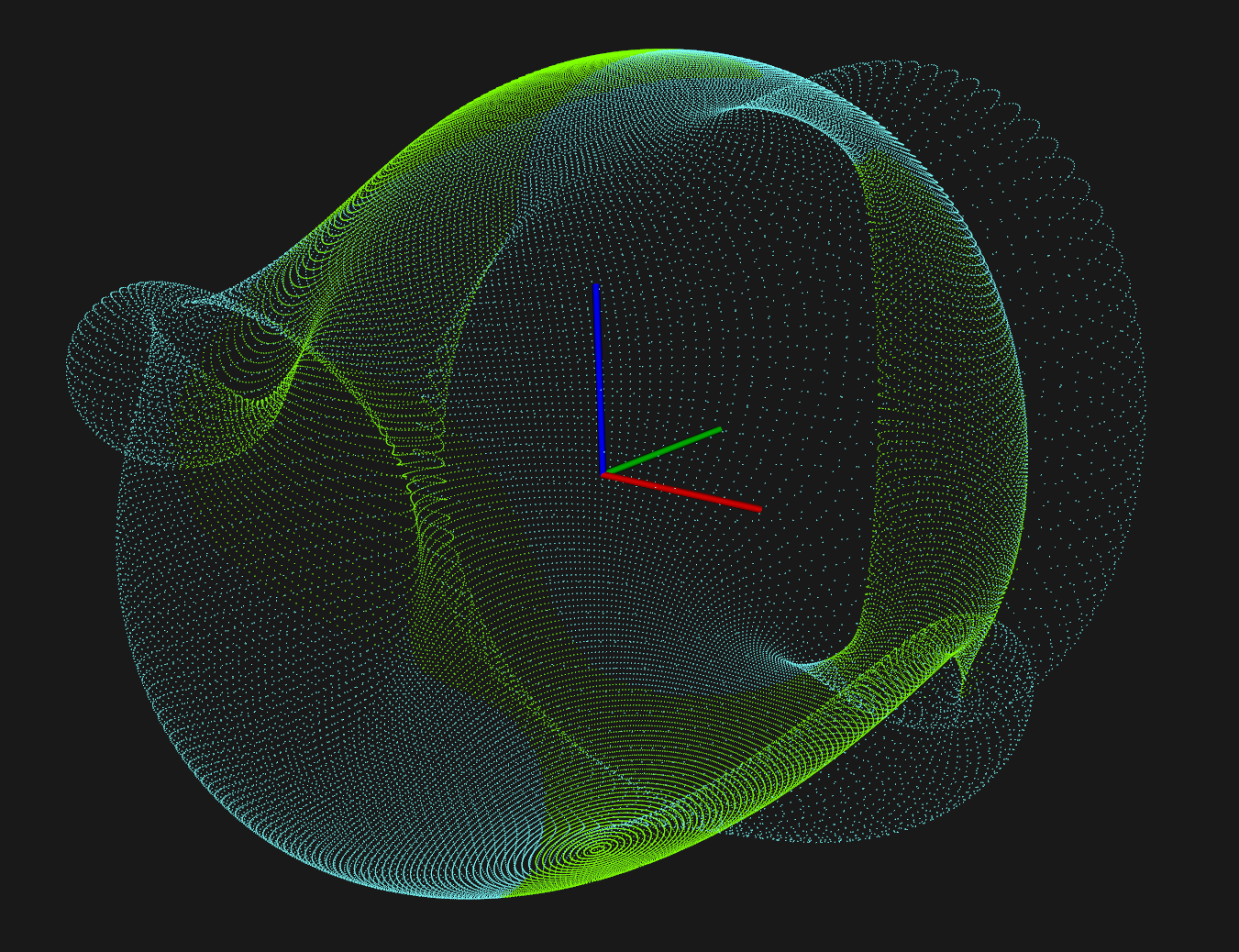}
    \caption{Simulation of magnetic field generated by three magnets}
    \label{fig_hw:magnet_viz}
\end{figure}

Figure~\ref{fig_hw} shows the hardware design of the ball-socket joint of Roboy 3.0. On the left image, the ball-in-socket includes two main components, the ball and the socket. Inside the 3D-printed socket is the PCB with four 3D-magnetic sensors TLE493d mounted around the circuit (Figure~\ref{fig_hw:2}). The sensor data is automatically read out by the Terasic DE10-Nano FPGA dev kit using an I2C switch TCA9546A. 
The sensors measure the magnetic field strength generated by three 10mm neodymium cube magnets mounted inside the ball. The magnets are distributed as depicted in Figure~\ref{fig_hw:magnet}. The magnetic field was simulated using magpylib \cite{magpylib2020} and the result is shown in Figure~\ref{fig_hw:magnet_viz}. The main goal of the design is to have as little symmetry in the magnetic field as possible, otherwise the same magnetic field would be sensed for different joint positions and such a function is impossible to learn with standard feed forward neural networks. Many magnet arrangements were evaluated both in simulation and hardware. The three magnet arrangement was found to provide sufficient sensor signal strength and little symmetry for our target joint angle range.

%% file: sections/03_Methodologies.tex
\subsection{Problem statement}

In this research, our objective is obtaining the orientation of the ball-and-socket joint based on the output of the magnetic sensors. Formally speaking, let $x_i \in \mathcal{X}$ be the magnetic output of the sensor $i$-th, $y \in SO(3)$ is our actual joint angle. We want to learn a transformation function $f$ such that:
\begin{align*}
    f(\{x\}_{i=1}^n) = y
\end{align*}
The problem seems to be easily solvable with a simple feedforward neural network. However, there are two problems:
\begin{itemize}
    \item Output of a feedforward neural network often lies in Euclidean space, i.e $\mathrm{R}^3$. Without further care, the output will be discontinuous whenever it approaches the singularity point. To solve this, we use a solution proposed by researchers from the computer graphics community \cite{zhou2018onthecontinuity}. In short, the output still lies in $\mathrm{R}^3$, but then will be transformed to the 6D vector representation which is possible to convert to the Euler angle afterwards. The proposed transformation is differentiable; hence the end-to-end learning fashion remains.
    \item The input source $x_i$ is noisy, and sometimes returns an outlier value. This leads to a undesirable consequence as the distorted output of the network will be propagated through the kinematic control stack, causing instabilities and unpredicted behavior. We therefore propose to treat the problem as the sequence model regression problem. Instead of relying on the only recent input, we take the advantage of having a time-series data, i.e series of input and expect that the output will be smooth by those non-corrupted historical input afterwards. Two proposed models will be explained more in detail in the following subsections.
\end{itemize}

\subsection{Recurrent Neural Network}

A Recurrent Neural Network (RNN) takes a sequence of inputs $(x_1, x_2, ...)$ and outputs a corresponding sequence $(y_1, y_2, ...)$. The network weights are shared time-wise. However, this naive design leads to a critical issue which is often referred to as "vanishing and exploding gradient" since after a certain time, the gradient becomes too small or too large, and the network will eventually stop learning. This well-known problem can be solved by the Long-short term memory (LSTM) \cite{HochSchm97} gate-mechanism. The main idea of LSTM is, instead of keeping all the information from history, we let the gates decide which information to keep, which to forget. In formal, let $C_{t-1}$ be the cell state at the previous time step, $\hat{C}_{t}$ be the updated state, and $C_{t}$ be the final state, the update will look like this:
\begin{align*}
    C_t = C_{t-1} \odot f_t + \hat{C}_t \odot i_t 
\end{align*}
where $f_t$ and $i_t$ are forget and input gate, respectively. Both are scalar, and if it is close to 0, $C_t$ will completely ignore the information from the previous cell $C_{t-1}$ and vice versa. Then, the final output can be obtained with this equation:
\begin{align*}
    y_t = o_t \odot tanh(C_t)
\end{align*}

\subsection{Deep Variational Bayes Filters}

A simple feedforward or recurrent (LSTM) model cannot provide a good estimate of the real-world dynamics and uncertainty. To mitigate this, we also try \emph{Fusion Deep Variational Bayes Filter} (Fusion DVBF) \cite{karl2017deep, karl2017unsupervised} in this research. In general, Fusion DVBF is a neural network that can perform filtering during inference and also has the possibility to fuse multiple sources of sensory input to obtain a more accurate state estimation (data fusion).

In terms of probabilistic sequential model, the conditional likelihood distribution is written as
\begin{align}
    p({x}_{1:T} \mid {u}_{1:T}) = \int p(x_1 | z_1)\rho(z_1) \prod_{t=2}^T p(x_t | z_t) p(z_t | z_{t-1}, u_{t-1}) dz_{1:T}
    \label{eq:dvbf_likelihood}
\end{align}
Maximizing Equation~\ref{eq:dvbf_likelihood} requires solving an integral, which is intractable as all the variables $x$, $z$, $u$ lie in the continuous space. One way to solve it is doing the maximization on the \emph{evidence lower bound} (ELBO) instead:

\begin{align*}
    \log{p({x}_{1:T}\mid {u}_{1:T})} 
    &\geq \mathcal{L}_{\text{ELBO}} ({x}_{1:T}, {u}_{1:T})  \\
    &= \mathrm{E}_q [\log{p({x}_{1:T}\mid {z}_{1:T}})] + \text{KL} (q({z}_{1:T}\mid {x}_{1:T}, {u}_{1:T})  \mid \mid p({{z}_{1:T}}))
\end{align*}

The $\mathcal{L}_{\text{ELBO}}$ is computed by introducing a variational distribution $q({z}_{1:T}\mid {x}_{1:T}, {u}_{1:T})$ to approximate the intractable $p({z}_{1:T}\mid {x}_{1:T}, {u}_{1:T})$. The main components of Fusion DVBF distinguish it from classical machine learning are:

\begin{enumerate}
    \item Amortized inference: the variational distribution $q$ was approximated by a neural network $q_\theta({z}_{1:T}\mid {x}_{1:T}, {u}_{1:T})$, which is also referred to as the recognition network.
    \item The recognition network can be written in a form that is similar to the classical Bayesian filter \cite{karl2017unsupervised}
    \begin{align}
        q(z_t | z_{1:t-1}, x_{1:t}, u_{1:t}) \propto q_{\text{meas}} (z_t | x_t) \times q_{\text{trans}} (z_t | z_{t-1}, u_t)
        \label{eq:dvbf_fusion}
    \end{align}
\end{enumerate}

These modifications form an end-to-end learning, in the style of \emph{variational auto-encoder} \cite{kingma2014autoencoding}. Moreover, modelling the inference network like Equation~\ref{eq:dvbf_fusion} allows us to put more sensory input knowledge via $q_{\text{meas}} (z_t | x_t)$ as in classical data fusion.

To apply Fusion DVBF into our problem, we need two further modifications. First, the original model only returns the latent representation of the input space. However, we also need a corresponding output (joint angle) from the sensory input (magnetic output). Therefore, we introduce another "decoder" network that maps the latent state $z$ to the output $y$, i.e $q(y \mid z)$. Second, although applying data fusion for multiple sensory input is straightforward, without any constraint, there is a high possibility that during training, the latent state results from each sensor will be far apart, which is non-intuitive for the aggregation afterwards. To solve this, we put a constraint that forces the latent states to stay close to each other. 

\subsubsection{Output-space transformation}

We introduce another random variable $y$ that will be stacked together with the observable $x$. This is our desired output. Together, they form the log likelihood distribution $\log(x_{1:T}, y_{1:T} \mid u_{1:T})$, with the corresponding ELBO

\begin{multline}
    \mathcal{L}_{\text{ELBO}} ({x}_{1:T}, {y}_{1:T}, {u}_{1:T}) = \mathrm{E}_q [\log{p({x}_{1:T}\mid {z}_{1:T}})] + \mathrm{E}_q [\log{p(y_{1:T} \mid z_{1:T})}] \\
    + \text{KL} (q({z}_{1:T}\mid {x}_{1:T}, {u}_{1:T})  \mid \mid p({{z}_{1:T}}))
\end{multline}

where $p(y \mid z)$ is another probabilistic neural network and acts as the decoder network like $p(x \mid z)$. 

\subsubsection{n-Sensors fusion}

We hypothesise that each sensor has enough information to infer the latent state. Therefore, we put them as an independent distribution, and together form the recognition distribution in a factorisation fashion:
\begin{align*}
    q_{\text{meas}}(\cdot) = \prod_{i=1}^n q_i (\cdot) =  \prod_{i=1}^n \mathcal{N}(\mu_i, \Sigma_i)    
\end{align*}
with $n$ is the number of sensors and the way the aggregated Gaussian parameters ($\mu_\text{meas}, \sigma_{meas}$) computed is as follow:

\begin{align*}
    \sigma_\text{meas} = \left ( \sum_{i=1}^n \frac{1}{\sigma_i^{2}} \right )^{-1/2}
\end{align*}

\begin{align*}
    \mu_\text{meas} = \sigma_\text{meas}  \sum_{i=1}^n {\sigma_i}^{-2} \times \mu_i
\end{align*}

Normally, due to the suboptimality, each $q_i(z | x, u)$ can have different value and still result in the same $\mu_\text{meas}$ and $\sigma_\text{meas}$. This is actually bad as one of the advantages of sensor fusion is that, in the aggregation step, it will penalise the less accurate sensor based on its variance. Therefore, it makes sense if the latent state resulting from every sensor stays close to each other. Then, even though there is corruption in the sensory inputs, it will be strongly penalised and consequently, other non-corrupted inputs can help to cover this. The extra soft constraint we put is:
\begin{align*}
    \mathcal{L}_{\text{ELBO}}^{\text{new}} ({x}_{1:T}, {y}_{1:T}, {u}_{1:T}) = \mathcal{L}_{\text{ELBO}} ({x}_{1:T}, {y}_{1:T}, {u}_{1:T}) + \alpha \| \mu_i (x) \|_2^2
\end{align*}
It simply forces the output of the encoder to stay close to zero, and since we do not want it to be too restrictive, we put a small weight $\alpha$ on the constraint.

%% file: sections/04_Experiments.tex
\subsection{Experiment setup}

\subsubsection{Data collection}

To collect data, we use an open loop control and do random walks on the 3D space: we randomly sample a target joint and do the linear interpolation between that point and the current position, and move the arm towards that point. Data is recorded at every 20ms along the movement, including the magnetic sensors output, the joint targets, and the joint pose. The joint pose is our ground truth obtained using two HTC Vive trackers by mounting one tracker on the base link (torso) and one the moving link (e.g. upper arm) and calculating orientation between the two, which yields the orientation of the desired 3-DoF joint.

\subsubsection{Data Analysis}

\begin{figure}
    \centering
    \includegraphics[width=\linewidth]{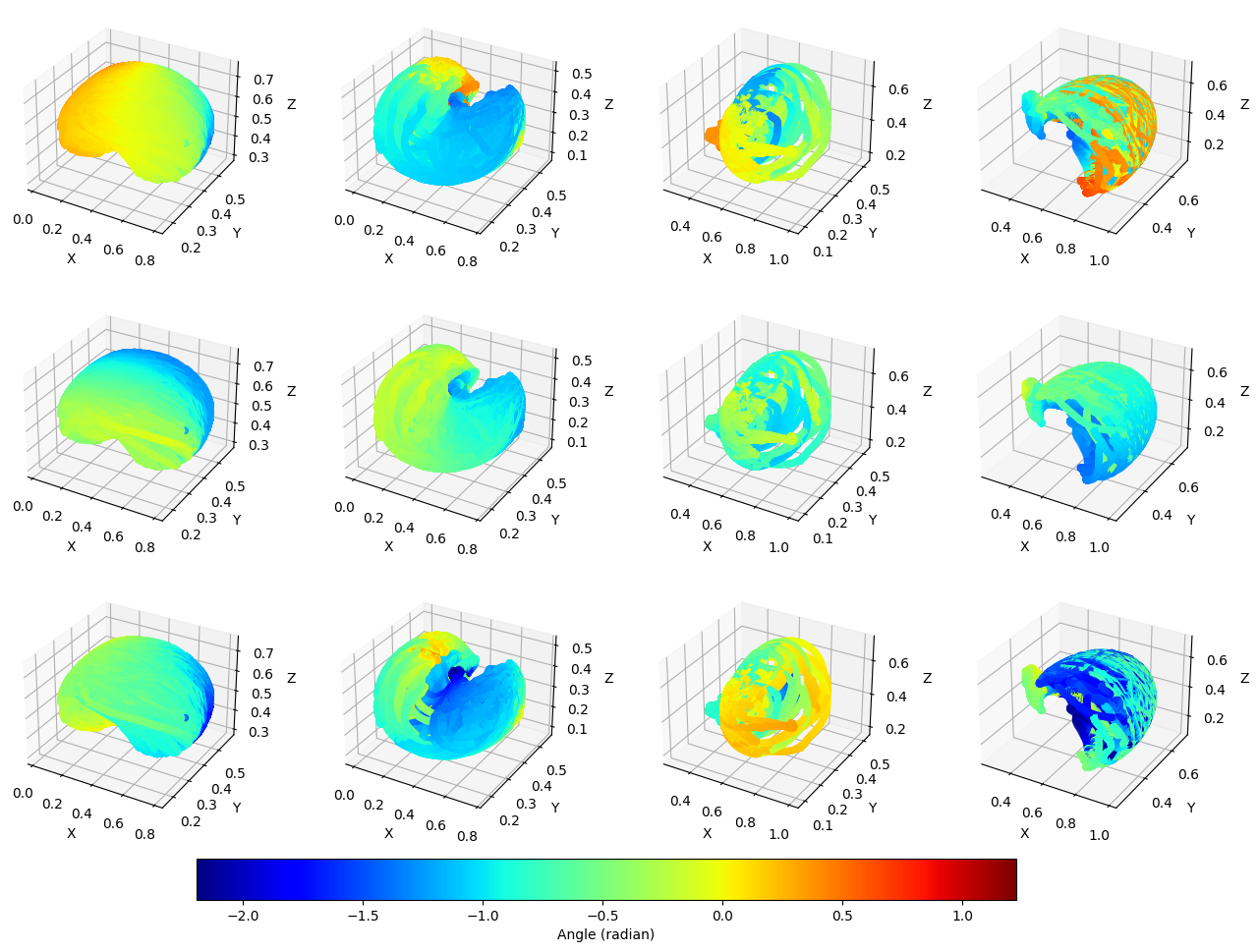}
    \caption{Correlation between outputs from magnetic sensor and the joint position. Columns are the magnetic sensors, and rows are the ground truth Euler angles.}
    \label{fig:data_plot}
\end{figure}

\begin{figure}
    \centering
    \includegraphics[width=\linewidth]{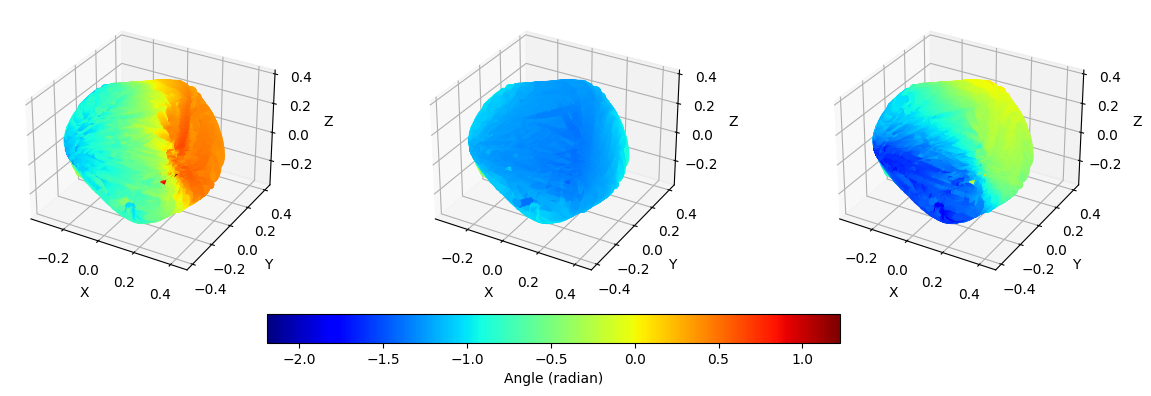}
    \caption{Correlation between outputs from magnetic sensor (reduced by Principal Component Analysis) and the joint position.}
    \label{fig:data_pca}
\end{figure}

The most basic assumption to be successful in the data driven approach is the correlation between inputs and outputs. If they are correlated, even weakly, there is hope that there exists a neural network that can approximate the actual function. Figure~\ref{fig:data_plot} and Figure~\ref{fig:data_pca} show this information, and in both figures, based on the color which indicates the output values (Euler angle), first, we do not see any clear discontinuities in the plots. This is important, otherwise learning would be very difficult or even impossible if the data are highly disconnected. On the first and second columns, the correlation looks really strong, we can see a lot of smooth curves, and together they even form a shape that looks like the entire workspace of the joint. 


\subsubsection{Evaluation}

To understand how well the proposed methods perform, we do the following setup on the left shoulder ball-and-socket joint of our robot:

\begin{itemize}
    \item For DVBF, our input is an 4D array of four magnetic sensor data, which will be treated individually by the architecture described in the previous section. DVBF acts as an online filtering method; data goes in and out one after another like a stream. We also feed to the DVBF the \emph{joint targets}, it acts as control input signal, and is of importance for the underlying state prediction phase.
    
    \item Unlike DVBF, in the LSTM network case, we only take five recent sensor outputs (no control input), feed them to the network and predict the next one. We follow this approach since LSTM is known to be sensitive with the historical data, and we do not want it to be too dependent on the far unuseful data.  
\end{itemize}

\subsection{Experiment results}

\begin{figure}
    \centering
    \includegraphics[width=0.7\linewidth]{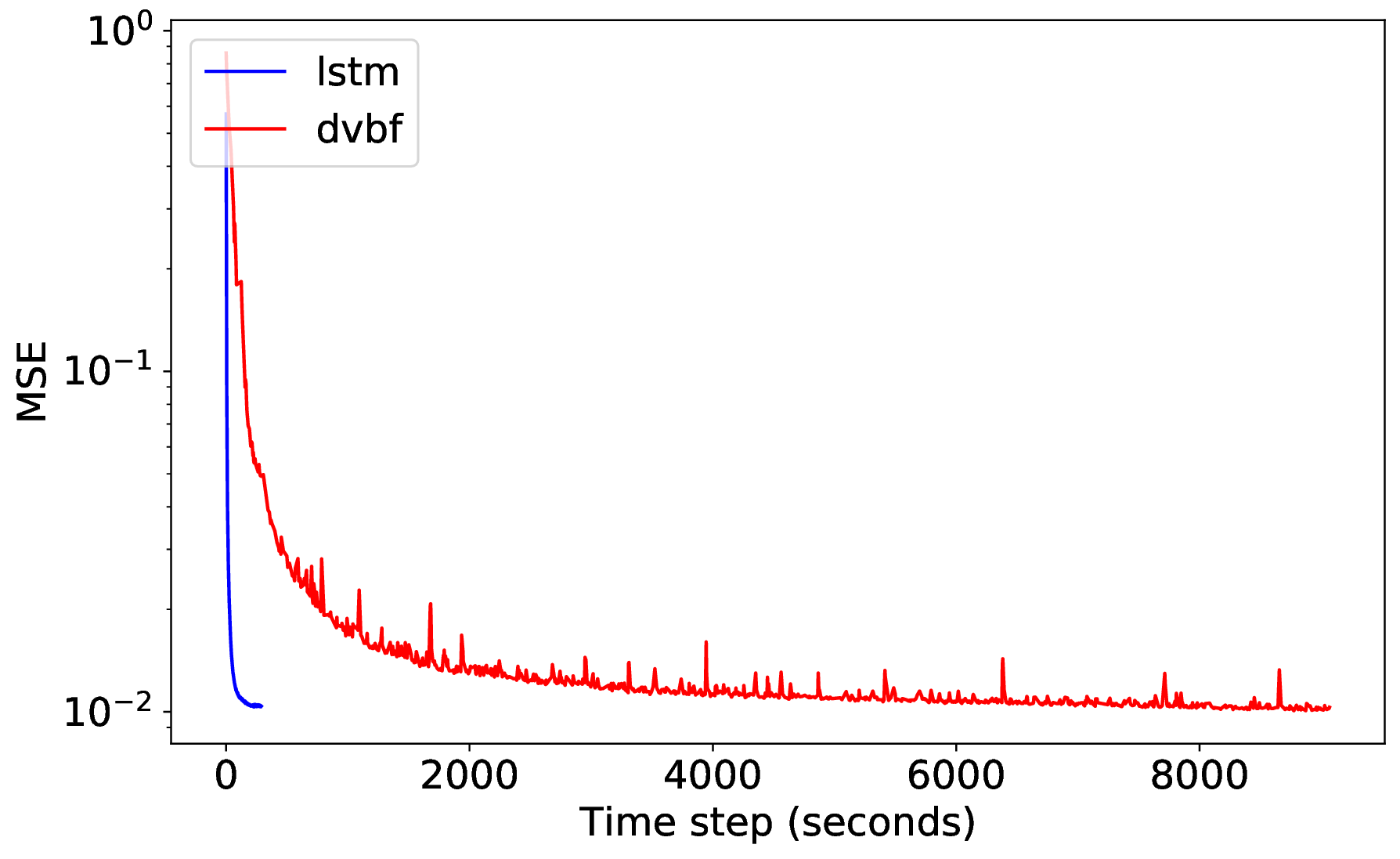}
    \caption{Validation MSE}
    \label{fig:validation_loss}
\end{figure}

\begin{figure}[ht] 

  \begin{subfigure}[b]{0.5\linewidth}
    \centering
    \includegraphics[width=\linewidth]{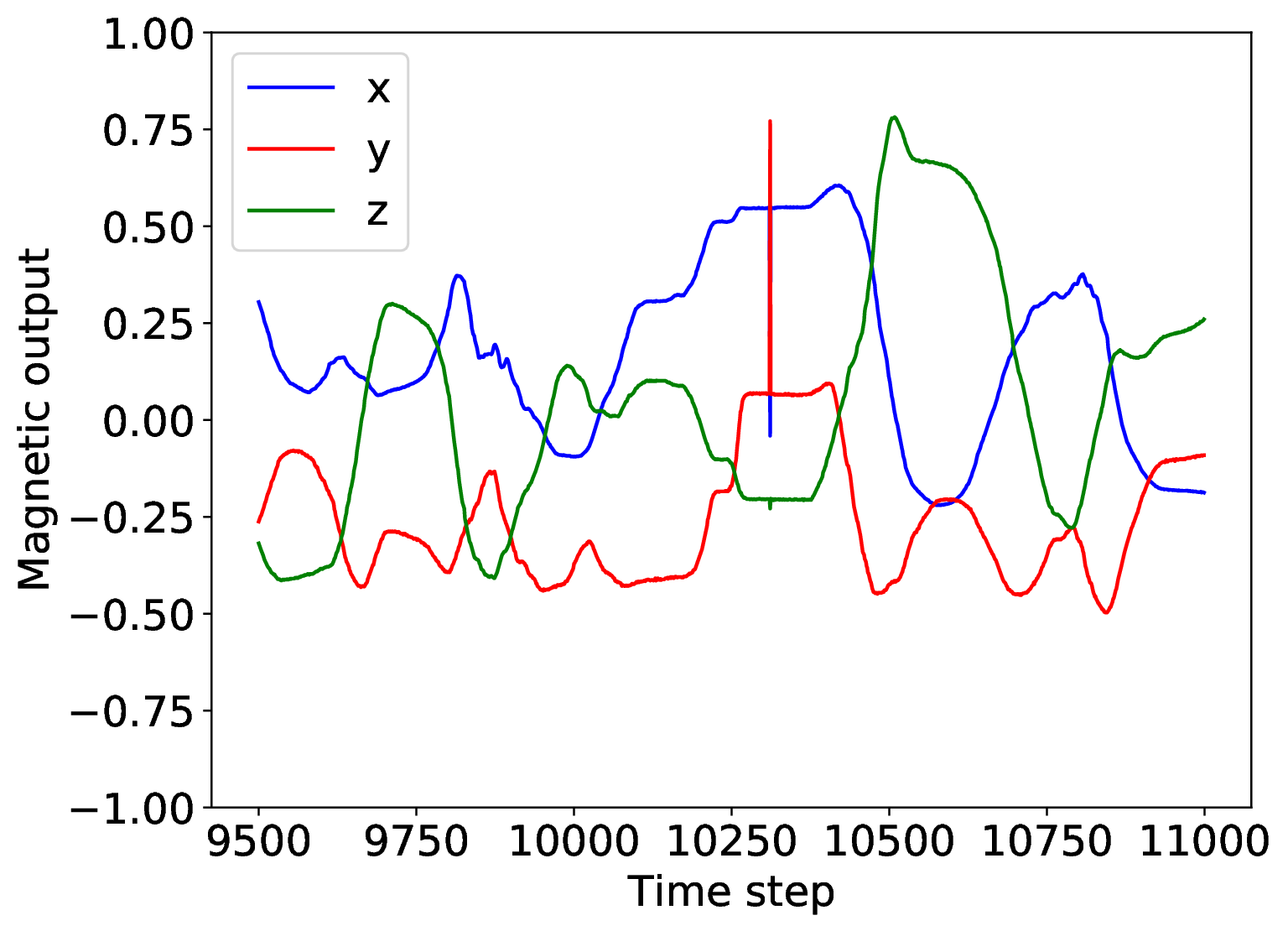}
    \caption{Magnetic sensor 1} 
    \label{fig:sensor_data:a} 
  \end{subfigure}
  \begin{subfigure}[b]{0.5\linewidth}
    \centering
    \includegraphics[width=\linewidth]{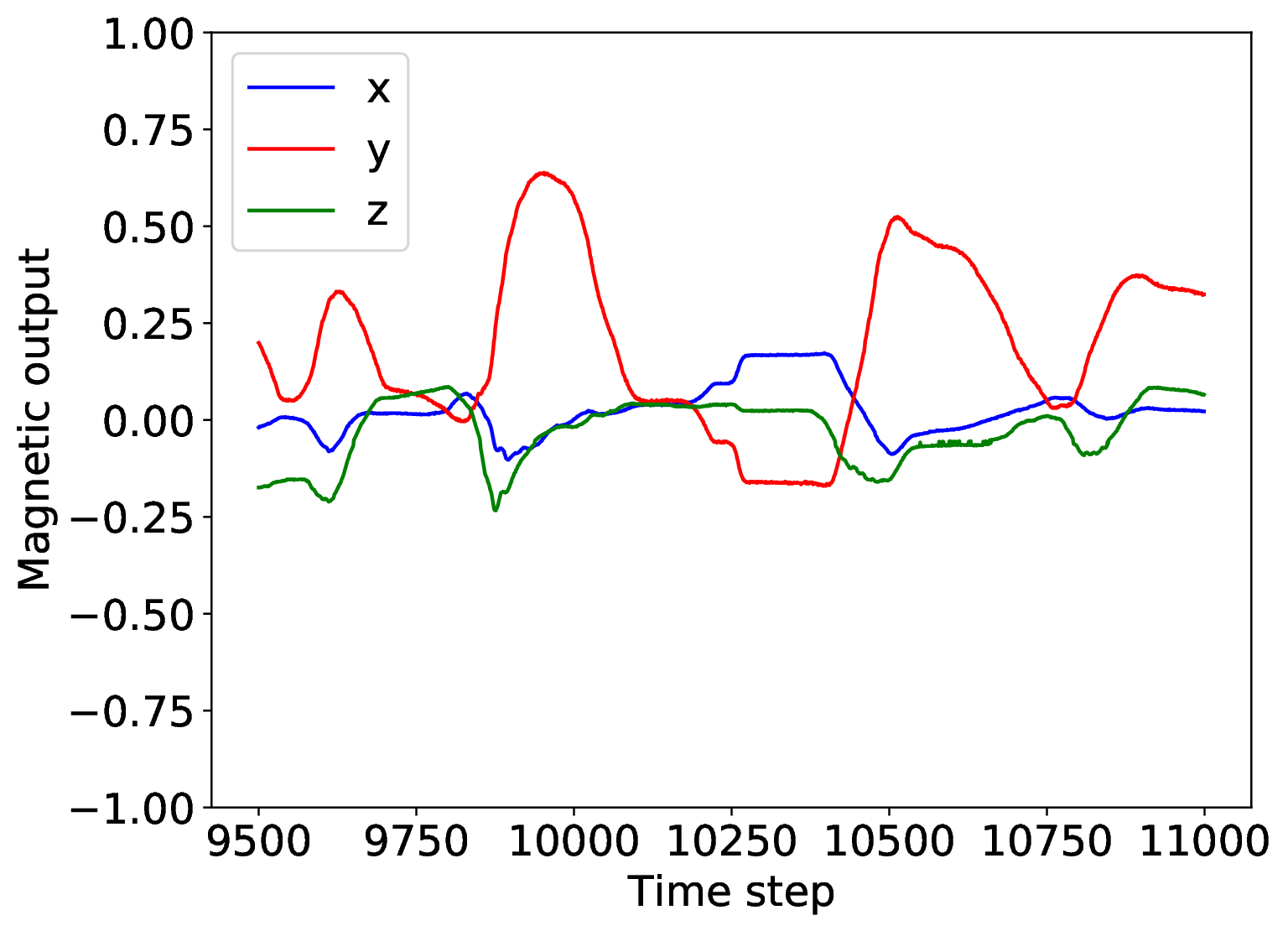}
    \caption{Magnetic sensor 2} 
    \label{fig:sensor_data:b} 
  \end{subfigure} 
  
  \begin{subfigure}[b]{0.5\linewidth}
    \centering
    \includegraphics[width=\linewidth]{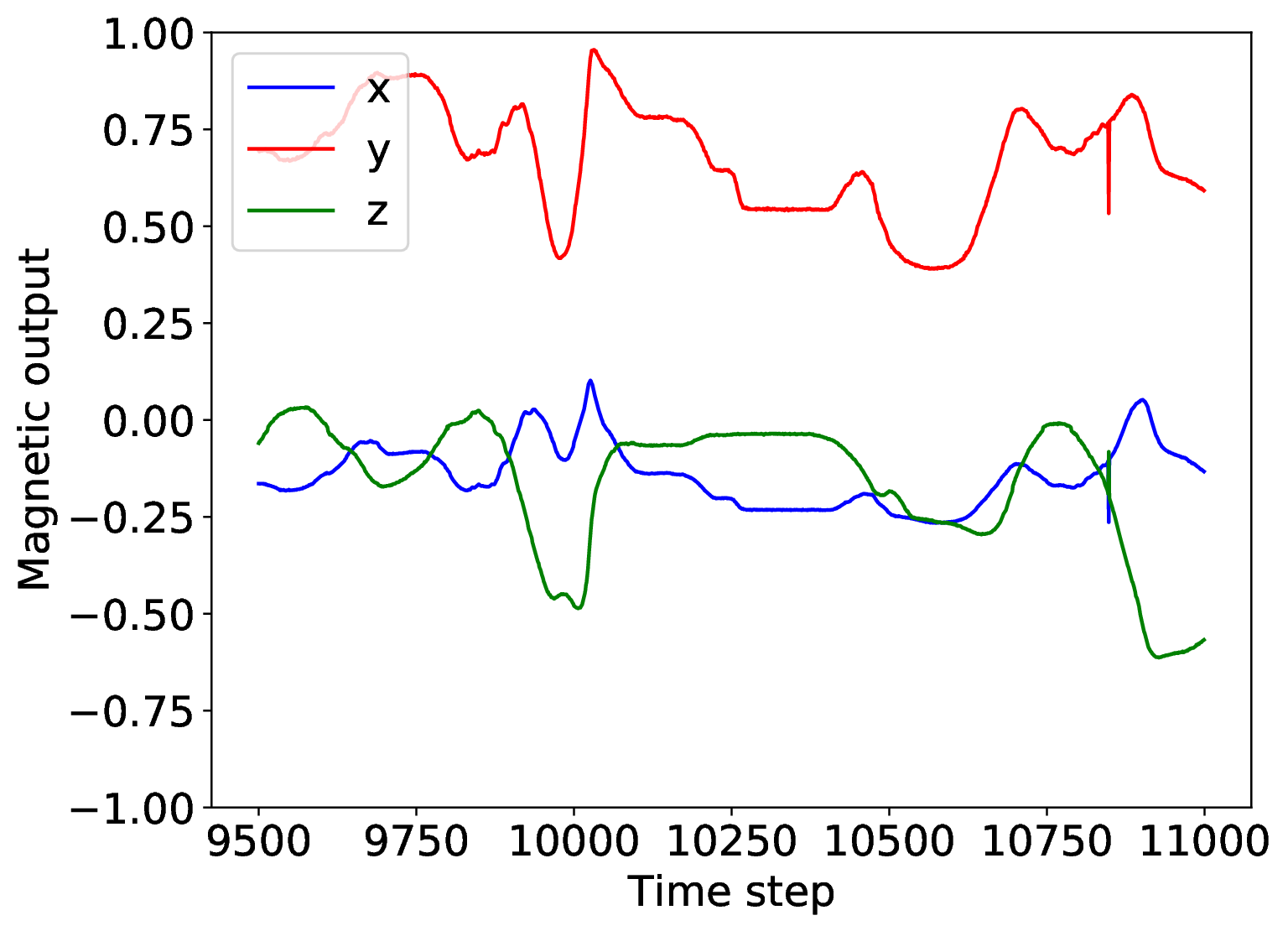}
    \caption{Magnetic sensor 3} 
    \label{fig:sensor_data:c} 
  \end{subfigure}
  \begin{subfigure}[b]{0.5\linewidth}
    \centering
    \includegraphics[width=\linewidth]{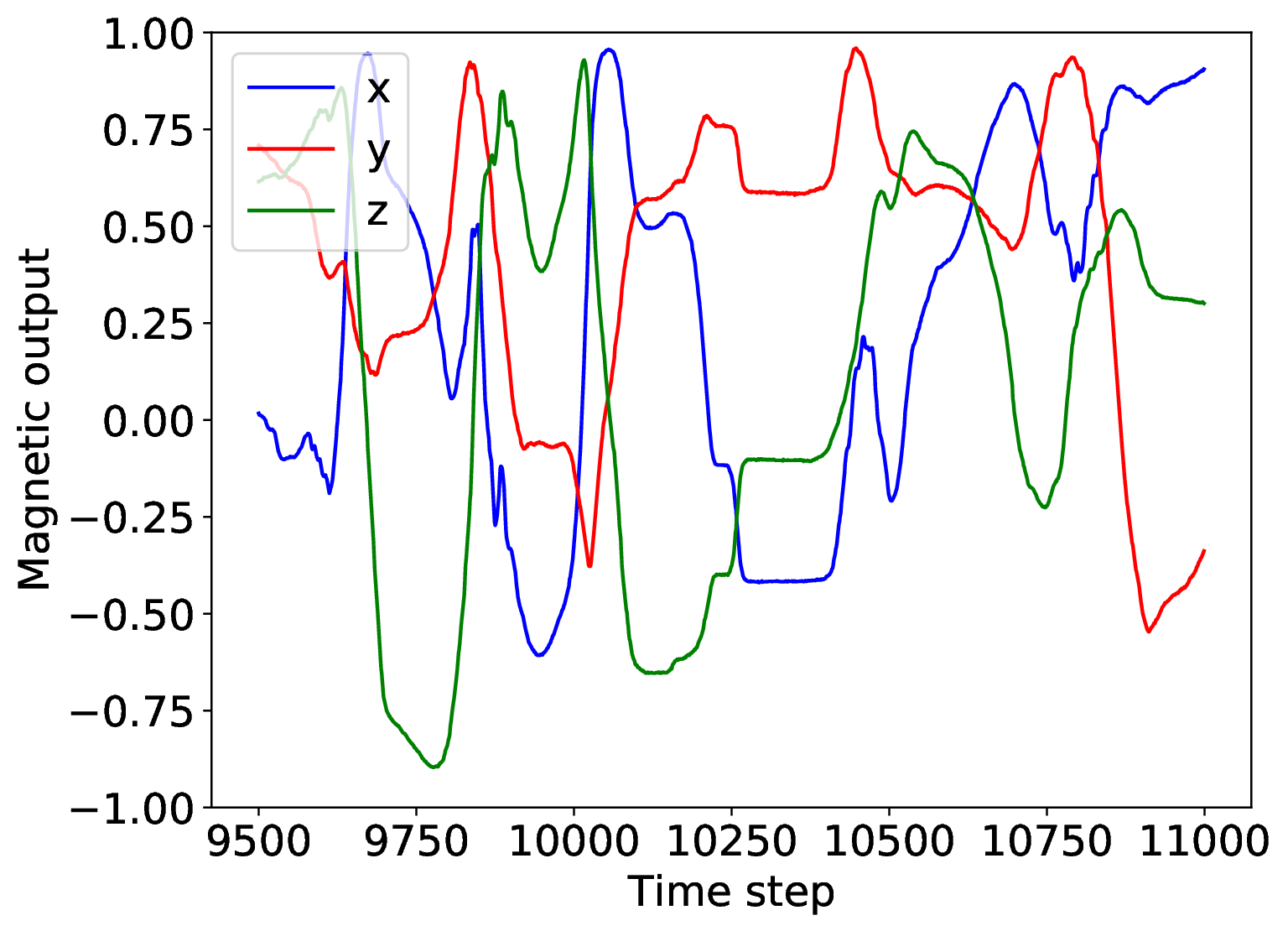}
    \caption{Magnetic sensor 4} 
    \label{fig:sensor_data:d} 
  \end{subfigure} 

  \caption{Magnetic sensor encoder values from time step 9500 to 11000 of the test dataset. There are four sensors placed around the disk, with the interval 20ms between two consecutive time steps}
  \label{fig:sensor_data} 
\end{figure}

First, we evaluate the training time of LSTM and DVBF. Based on Figure~\ref{fig:validation_loss}, it is obvious that to get the same mean square error (MSE) loss, DVBF needs significantly longer time, roughly two and a half hours compared to just 10 minutes from LSTM. This is quite surprising as we do not expect such a huge gap, probably the new networks introduced in the previous section makes DVBF struggling to find a good latent representation that satisfies all of the conditions.

We then demonstrate the performance of both the networks on the unseen data. To evaluate real-time accuracy of the proposed methods we executed a 3D random walk joint pose trajectory with the following characteristics:
\begin{itemize}
  \item trajectory duration: 404 s
  \item actual reached joint limits (in radians): 
  \begin{itemize}
      \item rotation around X axis: [-1.418,0.647]
      \item rotation around Y axis: [ -1.457, -0.0288]
      \item rotation around Z axis: [ -2.036, 0.061]
  \end{itemize}
  \item magnetic sensor data published at 65 Hz per joint
\end{itemize}

The experiment yielded real-time joint pose estimation with both methods at 37 Hz with MSE of 0.0382 rad for DVBF and 0.0362 rad for LSTM.

\begin{figure}[ht] 

  \begin{subfigure}[b]{0.5\linewidth}
    \centering
    \includegraphics[width=\linewidth]{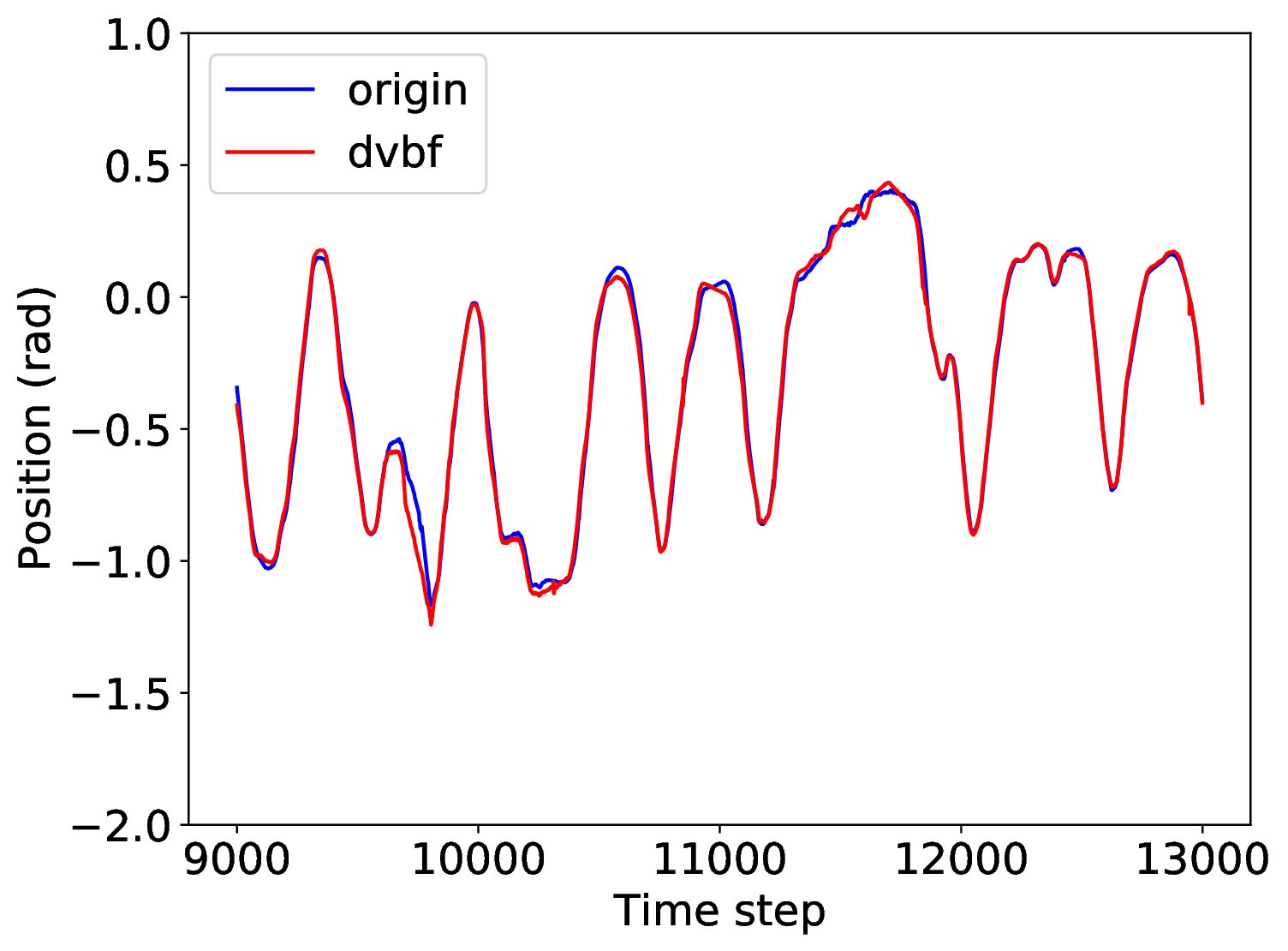}
    \caption{Euler X - DVBF} 
    \label{fig:comparison_euler:a} 
  \end{subfigure}
  \begin{subfigure}[b]{0.5\linewidth}
    \centering
    \includegraphics[width=\linewidth]{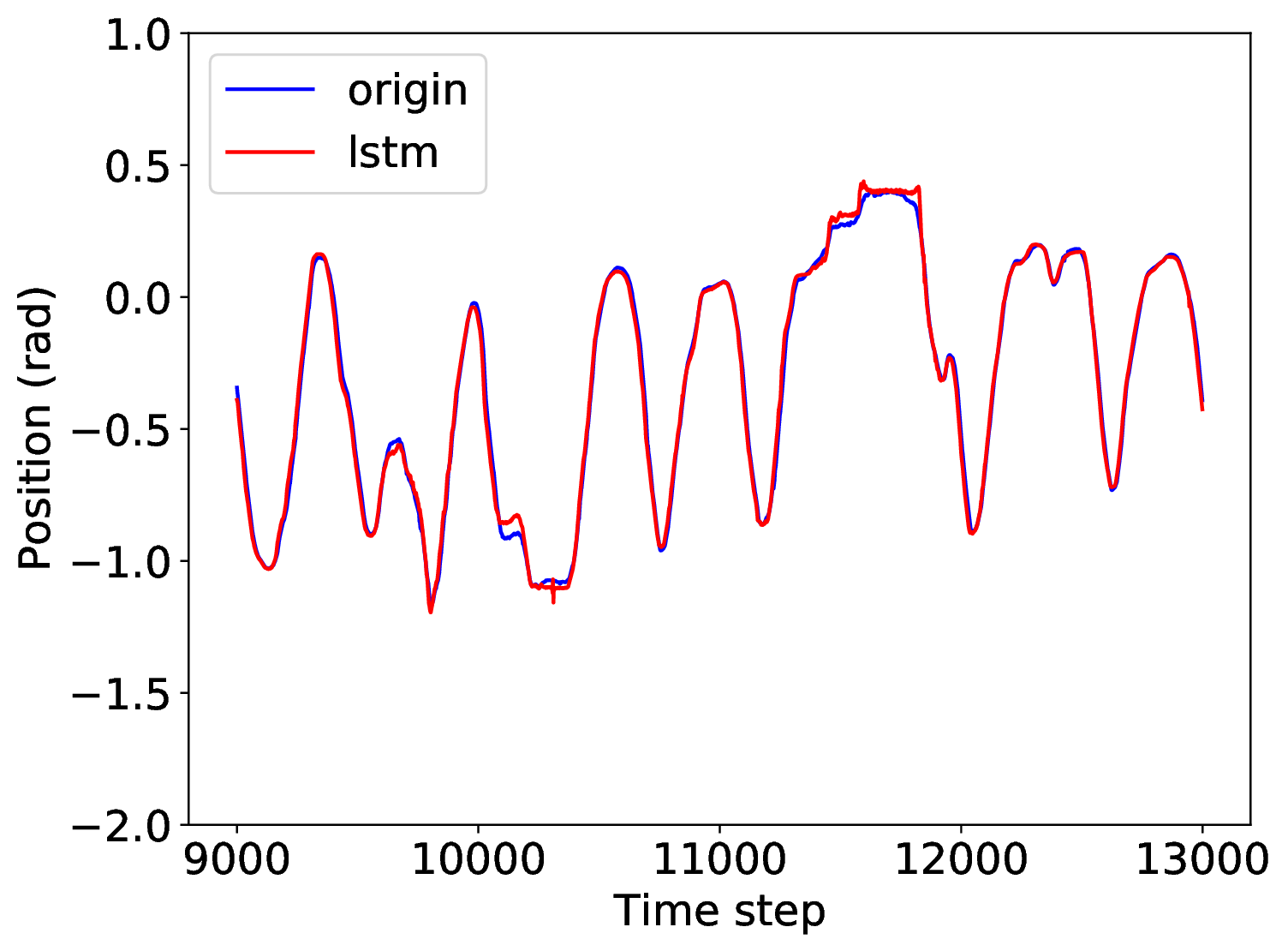}
    \caption{Euler X - LSTM} 
    \label{fig:comparison_euler:b} 
  \end{subfigure} 
  
  \begin{subfigure}[b]{0.5\linewidth}
    \centering
    \includegraphics[width=\linewidth]{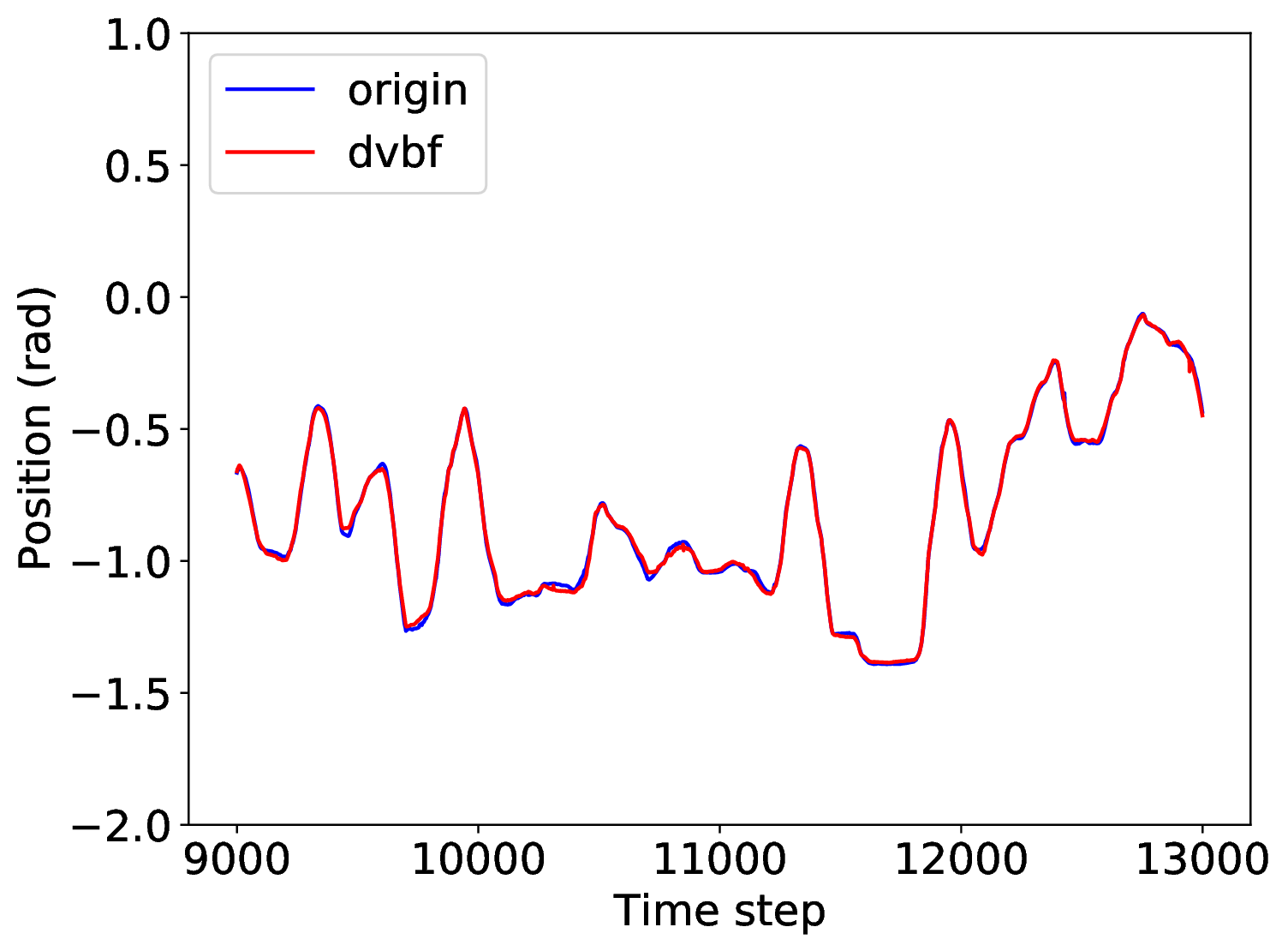}
    \caption{Euler Y - DVBF} 
    \label{fig:comparison_euler:c} 
  \end{subfigure}
  \begin{subfigure}[b]{0.5\linewidth}
    \centering
    \includegraphics[width=\linewidth]{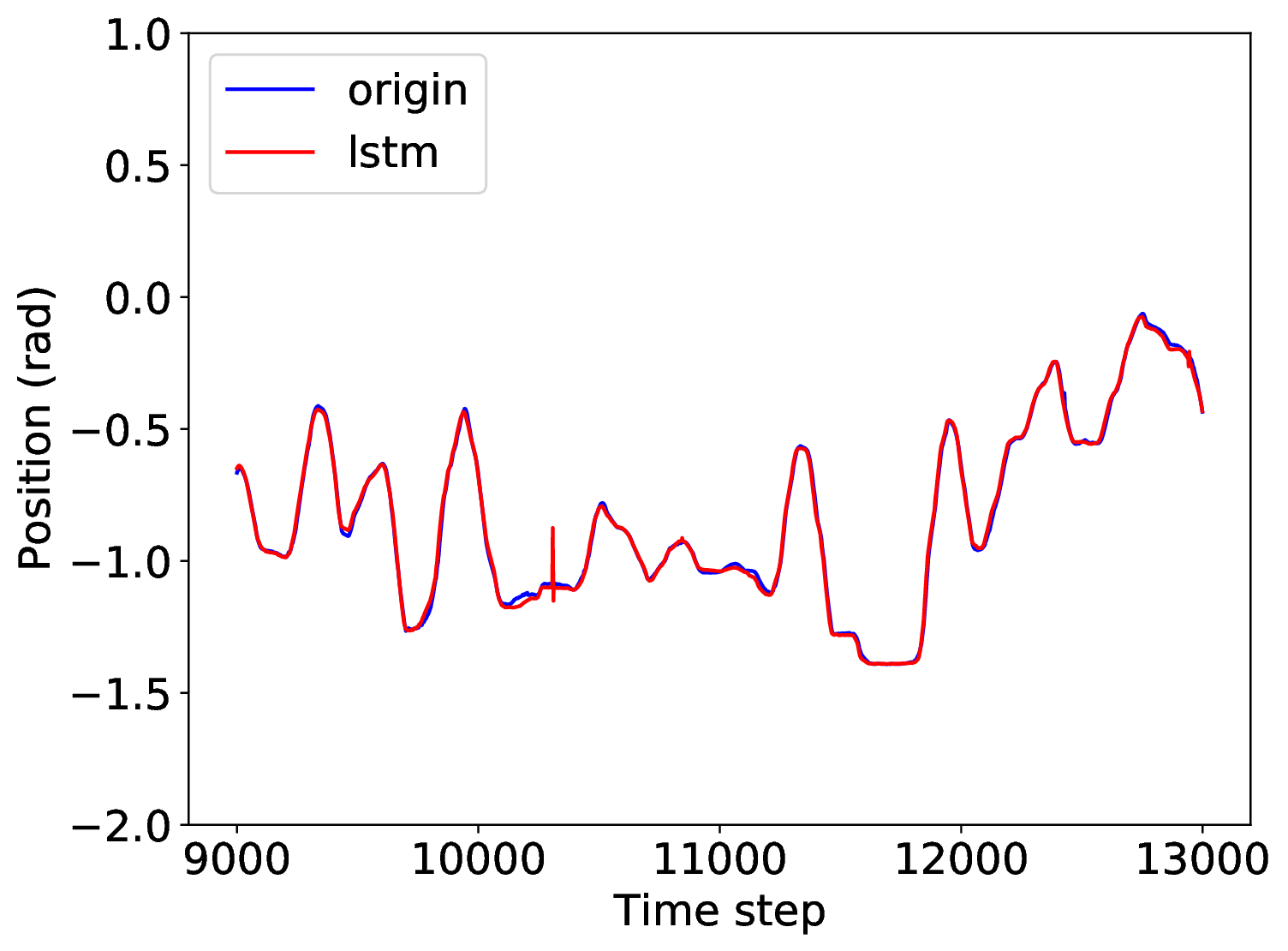}
    \caption{Euler Y - LSTM} 
    \label{fig:comparison_euler:d} 
  \end{subfigure} 
  
  \begin{subfigure}[b]{0.5\linewidth}
    \centering
    \includegraphics[width=\linewidth]{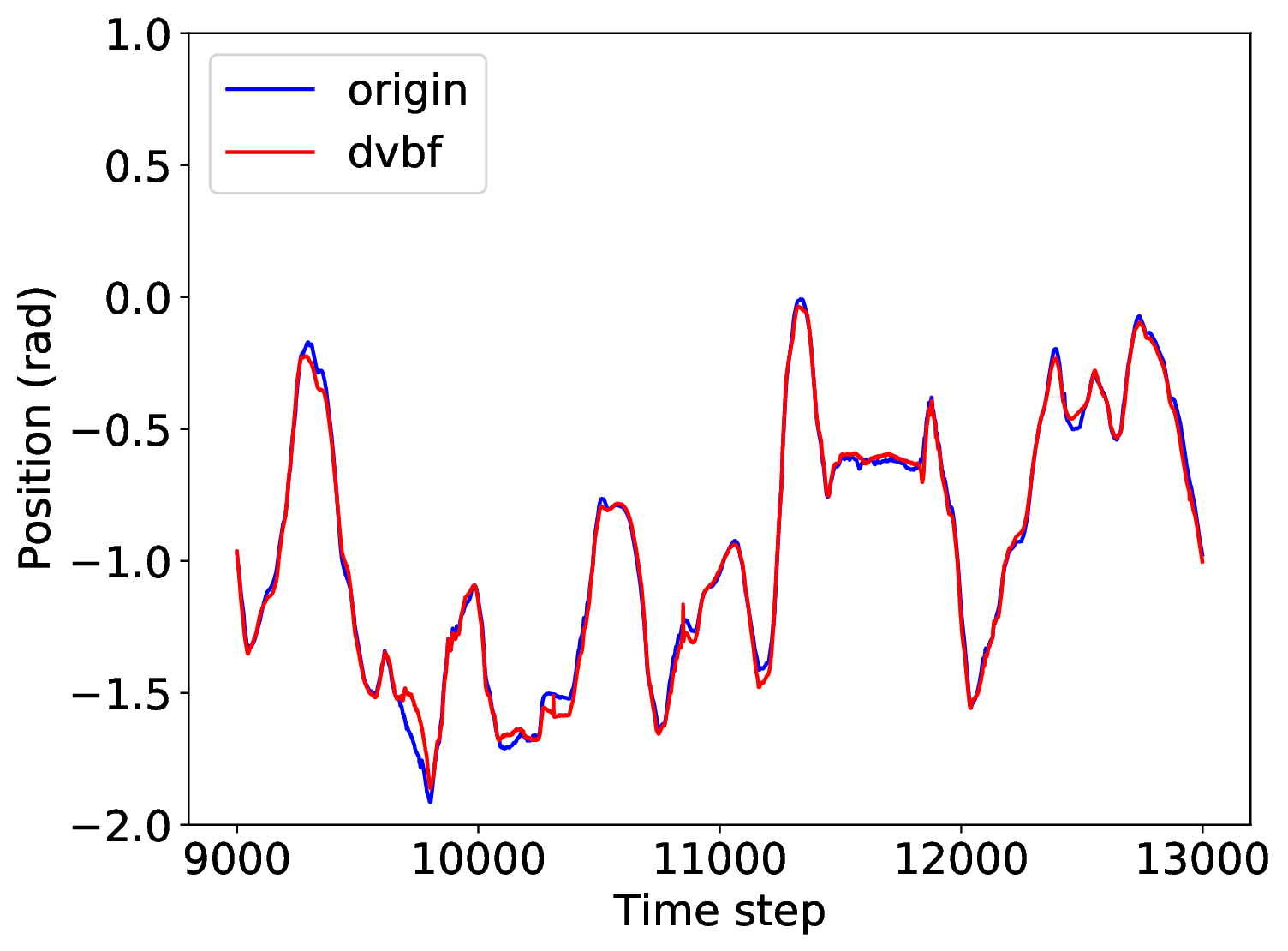}
    \caption{Euler Z - DVBF} 
    \label{fig:comparison_euler:e} 
  \end{subfigure}
  \begin{subfigure}[b]{0.5\linewidth}
    \centering
    \includegraphics[width=\linewidth]{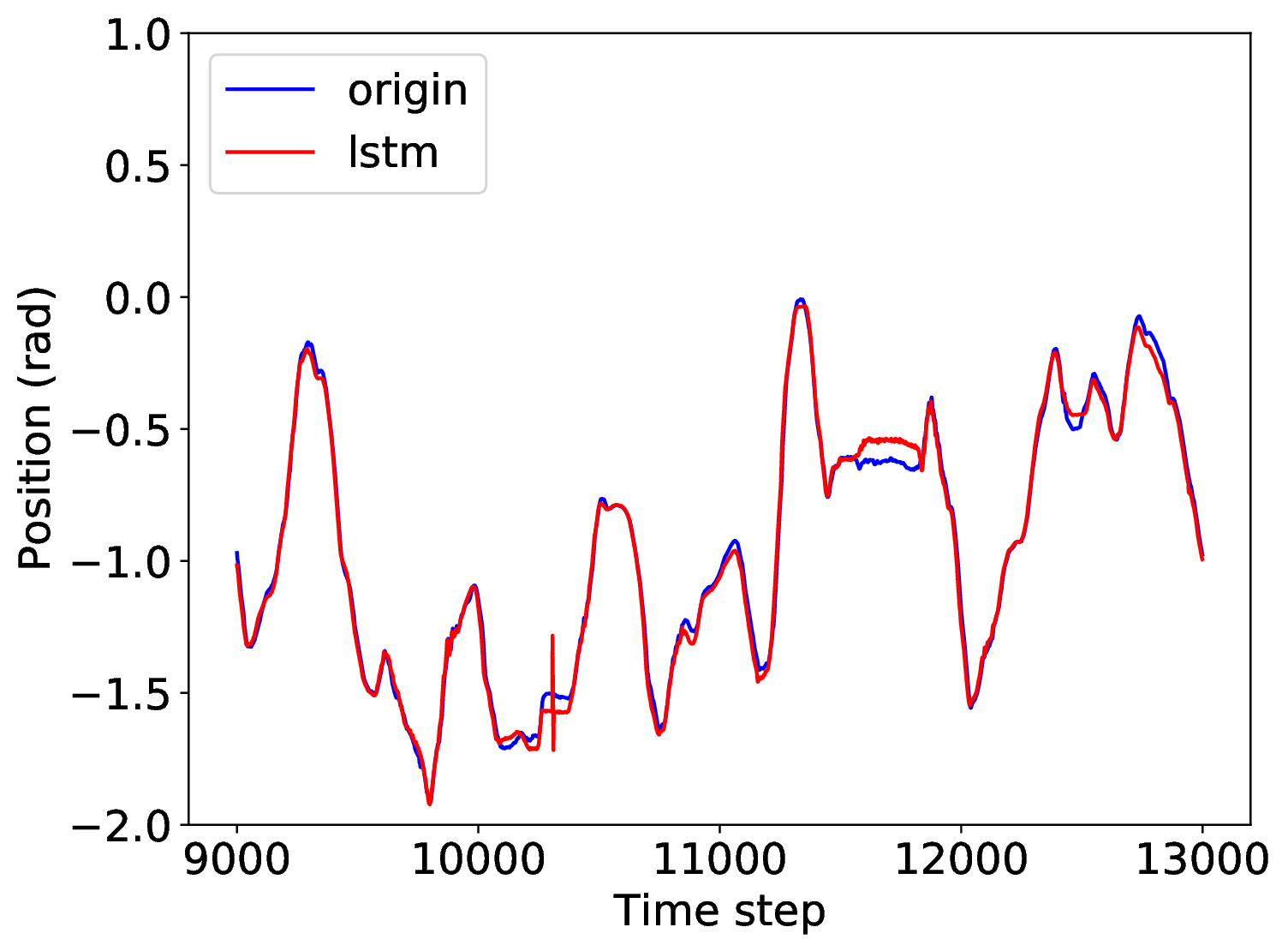}
    \caption{Euler Z - LSTM} 
    \label{fig:comparison_euler:f} 
  \end{subfigure} 
  
  \caption{Comparison between LSTM and DVBF with ground truth data from time step 9000 to 13000 of the test dataset. Data is plotted on three Euler angles, with the interval 20ms between two consecutive time steps}
  \label{fig:comparison_euler} 
\end{figure}

Further we analyze the quality of the predictions on a selected snapshot of our test data depicted on Figure~\ref{fig:comparison_error} and Figure~\ref{fig:comparison_euler}. Both LSTM and DVBF perform well on the period where there is no noise in the sensor data, the prediction output is indeed very close to the ground truth data. At around time step $10200$, however, there is a sudden jump in the magnetic output (Figure~\ref{fig:sensor_data:a}), and as a result, this sudden jump also appears in the prediction outputs on both of the methods. However, filtering-like DVBF is more advantageous in these cases, the learned state space model is able to recognize the corrupted sensors, and as a result, one can clearly see that the magnitude of the spike is much smaller compared to its corresponding output from LSTM model, though the jump is still recognizable from the Figure~\ref{fig:comparison_euler:e}. One can argue that those spikes can be eliminated by an appropriate post processing. However, we experienced that, it is only true if the jump magnitude is obviously an outlier. Otherwise, finding a compromise between the accuracy in the next prediction step and the appearance of an outlier is not that straightforward. This is the main reason why we choose DVBF, since this task is similar to the online filtering paradigm, finding the compromise between the model and the measurements.

\begin{figure}[ht] 

  \begin{subfigure}[b]{0.5\linewidth}
    \centering
    \includegraphics[width=\linewidth]{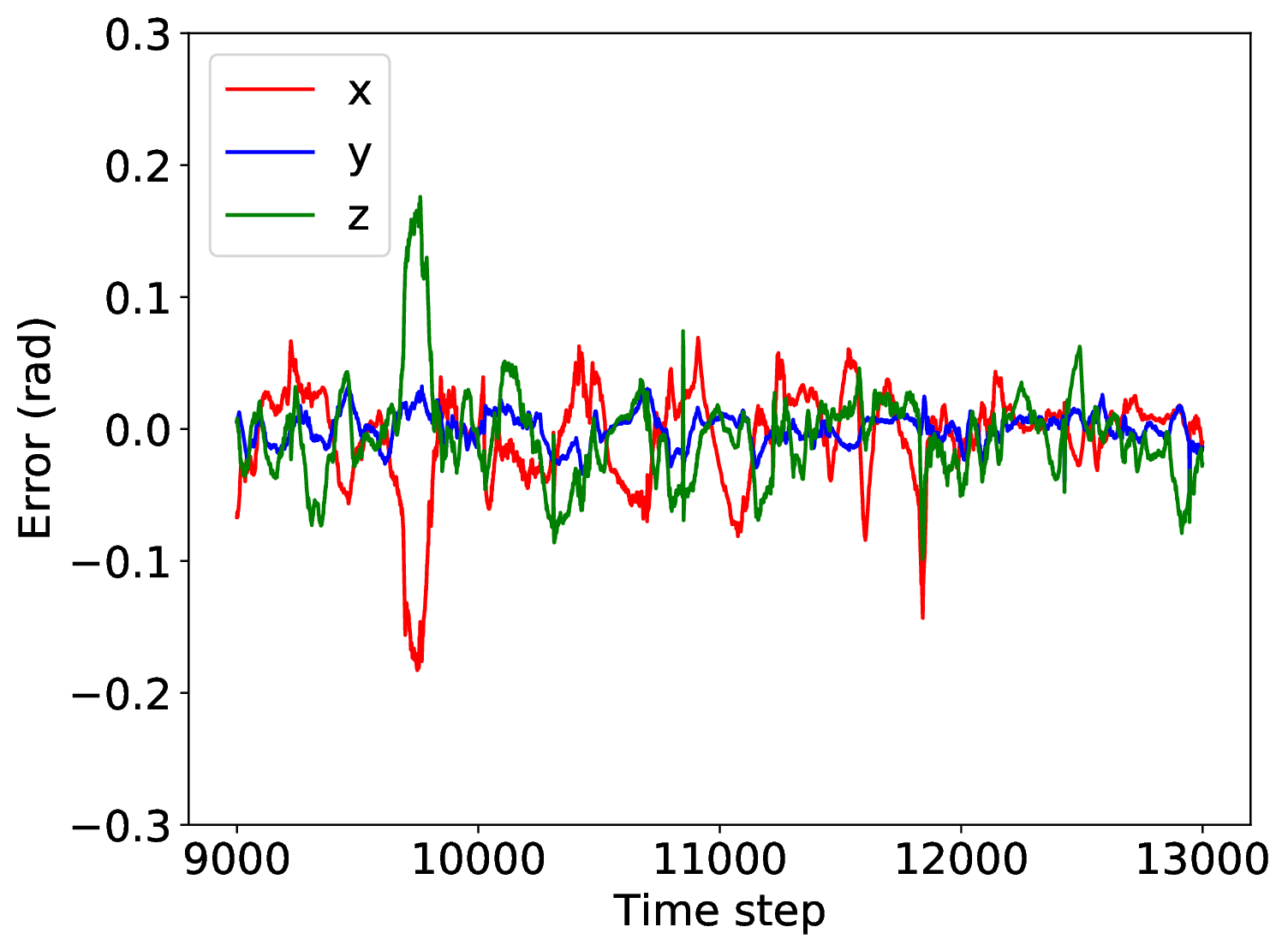}
    \caption{DVBF and ground truth} 
    \label{fig:comparison_error} 
  \end{subfigure}
  \begin{subfigure}[b]{0.5\linewidth}
    \centering
    \includegraphics[width=\linewidth]{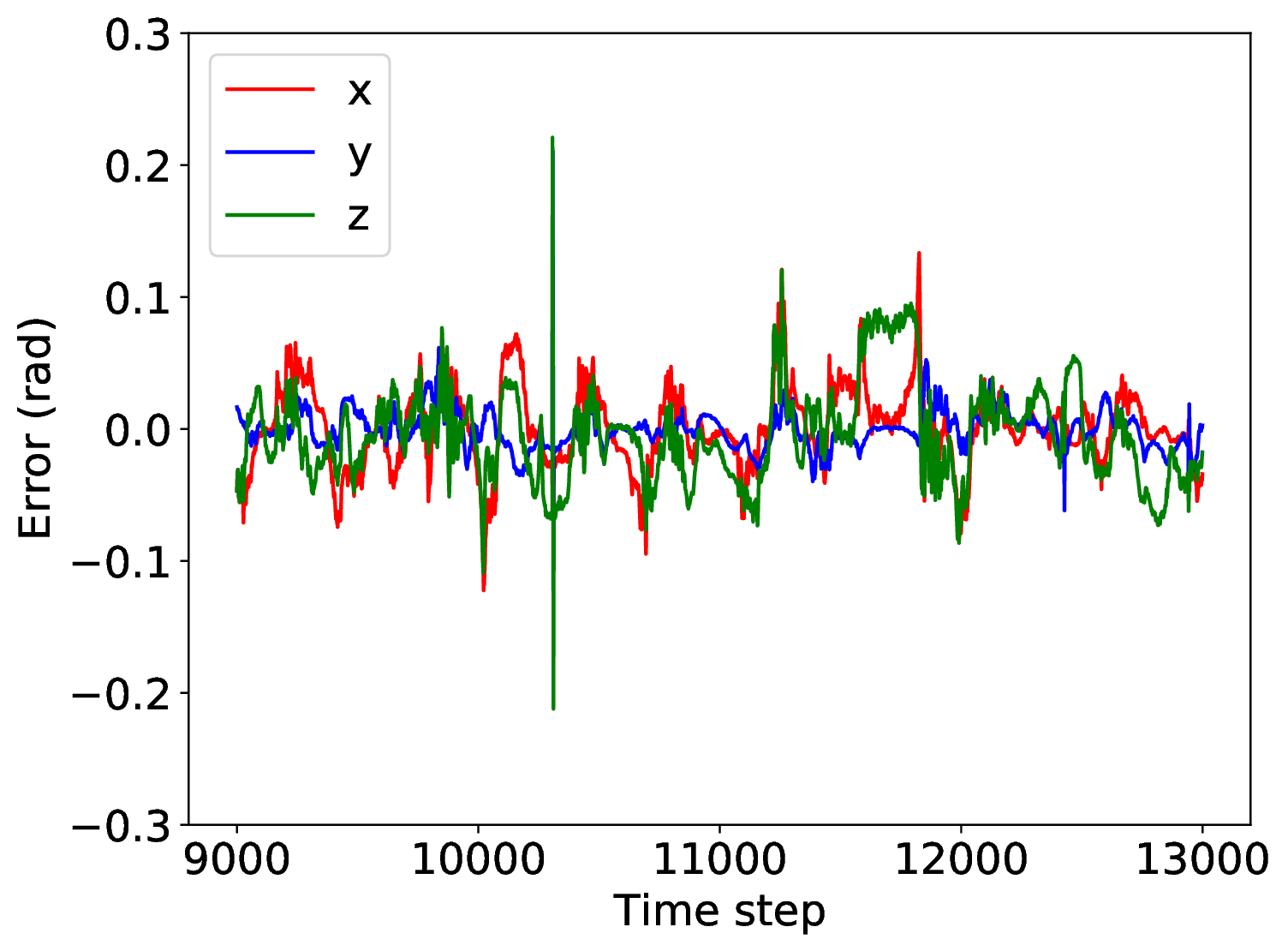}
    \caption{LSTM and ground truth} 
    \label{fig:comparison_error:b} 
  \end{subfigure}

  \caption{Error between our predictive model (DVBF and LSTM) and ground truth from time step 9000 to 13000 of the test dataset. Data is plotted on three Euler angles, with the interval 20ms between two consecutive time steps}
  \label{fig:comparison_error} 
\end{figure}

%% file: sections/05_Conclusion.tex
We presented in this paper a solution to estimate the orientation of the 3-DoF ball-and-socket joint. Magnetic field based approach was chosen due to its easy integrability and robustness of the environment. Although the magnetic field is stable, the three-dimensional Hall-effect sensor we used to get the magnetic value is known to be very noisy. We demonstrated in the paper that two sequential-based neural networks LSTM and DVBF can help mitigate this critical issue. The experimental results show that both LSTM and DVBF are able to successfully give a high accurate orientation of the joint. However, in the period when the magnetic sensor is corrupted, DVBF works better, the resulting output is smoother than LSTM method. Unfortunately, DVBF also has several clear disadvantages: the network structure is too sophisticated and tuning hyperparameters is not always a straightforward process despite potentially huge improvements in network's performance. Also, the training time of DVBF is significantly longer than LSTM. We therefore decided that, LSTM would be good solution for fast prototyping but then in the production phase, DVBF will be deployed to give a better estimation.

%% file: isrr22/sources.bib
@ARTICLE{roboy,  author={Richter, Christoph and Jentzsch, Soren and Hostettler, Rafael and Garrido, Jesus A. and Ros, Eduardo and Knoll, Alois and Rohrbein, Florian and van der Smagt, Patrick and Conradt, Jorg},  journal={IEEE Robotics   Automation Magazine},   title={Musculoskeletal Robots: Scalability in Neural Control},   year={2016},  volume={23},  number={4},  pages={128-137},  doi={10.1109/MRA.2016.2535081}}

@inproceedings{karl2017deep,
  author    = {Maximilian Karl and
               Maximilian Soelch and
               Justin Bayer and
               Patrick van der Smagt},
  title     = {Deep Variational Bayes Filters: Unsupervised Learning of State Space
               Models from Raw Data},
  booktitle = {5th International Conference on Learning Representations, {ICLR} 2017,
               Toulon, France, April 24-26, 2017, Conference Track Proceedings},
  publisher = {OpenReview.net},
  year      = {2017},
  url       = {https://openreview.net/forum?id=HyTqHL5xg},
  bibsource = {dblp computer science bibliography, https://dblp.org}
}

@ARTICLE{karl2017unsupervised,
      title={Unsupervised Real-Time Control through Variational Empowerment}, 
      author={Maximilian Karl and Maximilian Soelch and Philip Becker-Ehmck and Djalel Benbouzid and Patrick van der Smagt and Justin Bayer},
      year={2017},
      eprint={1710.05101},
      archivePrefix={arXiv},
      primaryClass={stat.ML}
}

@misc{kingma2014autoencoding,
      title={Auto-Encoding Variational Bayes}, 
      author={Diederik P Kingma and Max Welling},
      year={2014},
      eprint={1312.6114},
      archivePrefix={arXiv},
      primaryClass={stat.ML}
}

@misc{zhou2018onthecontinuity,
    Author = {Yi Zhou and Connelly Barnes and Jingwan Lu and Jimei Yang and Hao Li},
    Title = {On the Continuity of Rotation Representations in Neural Networks},
    Year = {2018},
    Eprint = {arXiv:1812.07035},
}

@Article{HochSchm97,
  author      = {Sepp Hochreiter and Jürgen Schmidhuber},
  journal     = {Neural Computation},
  title       = {Long Short-Term Memory},
  year        = {1997},
  number      = {8},
  pages       = {1735--1780},
  volume      = {9},
  optdoi      = {10.1162/neco.1997.9.8.1735},
  opteprint   = {http://dx.doi.org/10.1162/neco.1997.9.8.1735},
  opturl      = {http://dx.doi.org/10.1162/neco.1997.9.8.1735},
}

@INPROCEEDINGS{Junguk17,  
    author={Kim, Junguk and Son, Hungsun},  
    booktitle={2017 14th International Conference on Ubiquitous Robots and Ambient Intelligence (URAI)},   
    title={Two-DOF orientation measurement system for a magnet with single magnetic sensor and neural network},   
    year={2017},  
    volume={},  
    number={}, 
    pages={448-453},  
    doi={10.1109/URAI.2017.7992773}
}

@ARTICLE{Lee04,
  author={Kok-Meng Lee and Debao Zhou},
  journal={IEEE/ASME Transactions on Mechatronics}, 
  title={A real-time optical sensor for simultaneous measurement of three-DOF motions}, 
  year={2004},
  volume={9},
  number={3},
  pages={499-507},
  doi={10.1109/TMECH.2004.834642}
}

@INPROCEEDINGS{Garner01,
  author={Garner, H. and Klement, M. and Kok-Meng Lee},
  booktitle={2001 IEEE/ASME International Conference on Advanced Intelligent Mechatronics. Proceedings (Cat. No.01TH8556)}, 
  title={Design and analysis of an absolute non-contact orientation sensor for wrist motion control}, 
  year={2001},
  volume={1},
  number={},
  pages={69-74 vol.1},
  doi={10.1109/AIM.2001.936432}
}

@INPROCEEDINGS{Phil19,
  author={Meier, Phil and Rohrmann, Kris and Sandner, Marvin and Prochaska, Marcus},
  booktitle={2019 IEEE 1st Global Conference on Life Sciences and Technologies (LifeTech)}, 
  title={Application of magnetic field sensors for hand gesture recognition with neural networks}, 
  year={2019},
  volume={},
  number={},
  pages={200-203},
  doi={10.1109/LifeTech.2019.8884006}
}

@ARTICLE{Chiang20,
  author={Chiang, Ting-Hui and Sun, Zao-Hung and Shiu, Huan-Ruei and Lin, Kate Ching-Ju and Tseng, Yu-Chee},
  journal={IEEE Sensors Journal}, 
  title={Magnetic Field-Based Localization in Factories Using Neural Network With Robotic Sampling}, 
  year={2020},
  volume={20},
  number={21},
  pages={13110-13118},
  doi={10.1109/JSEN.2020.3003404}
}

@Article{Sasaki22,
AUTHOR = {Sasaki, Ai-ichiro},
TITLE = {Effectiveness of Artificial Neural Networks for Solving Inverse Problems in Magnetic Field-Based Localization},
JOURNAL = {Sensors},
VOLUME = {22},
YEAR = {2022},
NUMBER = {6},
ARTICLE-NUMBER = {2240},
URL = {https://www.mdpi.com/1424-8220/22/6/2240},
PubMedID = {35336410},
ISSN = {1424-8220},
DOI = {10.3390/s22062240}
}

@Article{magpylib2020,
title = {Magpylib: A free Python package for magnetic field computation},
author  ={Ortner, Michael and Coliado Bandeira, Lucas Gabriel},
year    = {2020},
journal = {SoftwareX},
publisher = {Elsevier},
doi = {10.1016/j.softx.2020.100466}
}

@INPROCEEDINGS{cardsflow,
  author={Trendel, Simon and Chan, Yin Pok and Kharchenko, Alona and Hostetrler, Rafael and Knoll, Alois and Lau, Darwin},
  booktitle={2018 IEEE-RAS 18th International Conference on Humanoid Robots (Humanoids)}, 
  title={CARDSFlow: An End-to-End Open-Source Physics Environment for the Design, Simulation and Control of Musculoskeletal Robots}, 
  year={2018},
  volume={},
  number={},
  pages={245-250},
  doi={10.1109/HUMANOIDS.2018.8624940}}
